# Dendritic Cells for Real-Time Anomaly Detection

Julie Greensmith[*] and Uwe Aickelin[*]

[*]School of Computer Science, University of Nottingham, UK

`jqg, uxa@cs.nott.ac.uk`

## Abstract

Dendritic Cells (DCs) are innate immune system cells which have the power to activate or suppress the immune system. The behaviour of human DCs is abstracted to form an algorithm suitable for anomaly detection. We test this algorithm on the real-time problem of port scan detection. Our results show a significant difference in artificial DC behaviour for an outgoing portscan when compared to behaviour for normal processes.

## 1 Introduction

Intrusion detection systems (IDS) are a method used in computer security for detection of unauthorised use of machines. The Danger Project proposed by Aickelin et al. (2003) aims to improve on results previously seen with artificial immune systems (AIS) by applying concepts from the Danger Theory to IDS. Danger theory proposes that exposure to danger signals or pathogenic bacteria causes the activation of the immune system, not pattern matching of antigen. The cells responsible for combining these various signals are Dendritic cells. We use the 'signals plus context' processing power of Dendritic Cells (DCs) to perform anomaly detection.

Abstraction of certain properties thought key to DC function was performed, with algorithmic details and sources of biological inspiration detailed in Greensmith et al. (2005) . The properties we abstract from DC behaviour include their existence in different states, depending on their environmental conditions. As immature DCs, they collect multiple antigens and are exposed to signals, derived from dying cells in the tissue (safe or danger signals). DCs can combine these signals with bacterial signatures (PAMPs) to generate different output concentrations of costimulatory molecules, semi-mature cytokines and mature cytokines. Exposure to signals generates an increase in co-stimulatory molecules and causes the maturation of a DC to two different states: mature and semi-mature. DCs process a multitude of signals generated by the presence of bacteria or generated by damage to the tissue. PAMPs, based on a pre-defined signature, and danger signals (released on damage to the tissue) cause an increase in mature DC cytokines. Safe signals cause an increase in semi-mature DC cytokines and have a suppressive effect on both PAMPs and danger signals.

A key feature of the DCs is an ability to combine signals with antigen. In order to provide an environment suitable for the collection of signals and antigen, we use a system developed for the Danger Project (Aickelin et al. (2003)) known as **libtissue**. Using **libtissue** we can create a tissue compartment, to house a population of DCs. This compartment, known as a tissue server, is used to update the DCs on exposure to signals and antigen. A tissue client transforms raw values into normalised signal concentrations. A weighted signal processing function (described in detail in Greensmith et al. (2005)) is used to combine these signals to determine the output signal concentration of a DC. The exposure of a DC to PAMPs, danger or safe signals causes an increase in co-stimulatory molecules (CSM) on the DC. Once the CSM value exceeds a given threshold, the DC 'matures' and is removed from the sampling population. The concentrations of mature or semi-mature cytokines expressed by the DCs are calculated. Antigen is collected during the period of signal exposure in the sampling pool. Once a DC has matured, any antigen collected is labelled with presented DC context (mature or semi-mature) for each antigen. A mean percentage mature antigen value can be calculated, indicating the number of times an antigen was presented in a mature context.

## 2 Port Scan Experiment

For this experiment an ICMP (ping) scan is used to provide an example of malicious behaviour. System calls for a monitored secure shell (ssh) session are captured using a tissue client. This includes normal processes such as the controlling shell, x-forwarding agent and ssh demons. The process IDs of these system calls form the antigen. Signals are captured from different aspects of system behaviour. PAMPs are signature based and are derived from the number of ICMP errors received per second. Danger signals are derived from the number of packets per second sent by the machine. Safe signals are represented as the rate of change of packets per second, based over a 2 second moving average.

These incoming signals are used to convert DCs to either semi-mature or mature, measured by the relative concentrations of their output cytokines. For the duration of the experiment, the IDs of running processes, the output cytokines and the presented antigen are recorded. Post-hoc analysis allows us to measure the mean % mature context antigen for each process. Each experiment is repeated 10 times and an average value for the mean % mature context antigen is calculated, per process.

Three experiments are performed, using different combinations of signals and variations on the weight of the suppressive safe signal. Experiment 1 uses danger and safe signals alone, with a -1 value for the suppression by safe signals; experiment 2 uses danger and safe signals in combination with PAMPs; and finally, experiment 3 uses PAMPS, danger and safe signals, but with an increased value of suppression, a value of -2, to allow for exploration between this value and the detection of normal processes.

## 2.1 Results and Analysis

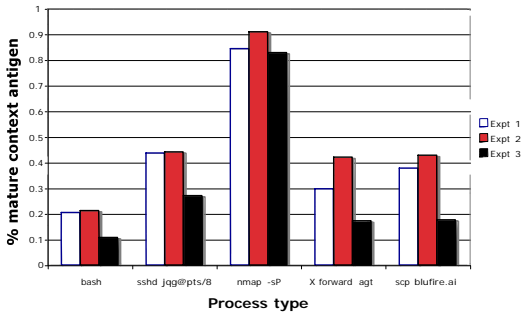

Figure 1: Experimental results, showing the % mature context antigen for each process

The five processes of interest presented in Figure 1 include: the bash shell from which the scan was performed; the ssh demon; the nmap performing the port scan; the graphical forwarding agent for the remote shell; and finally, the file transfer (scp). The detection of the nmap was significantly greater than the value derived for the normal file transfer, especially in experiment 3. The addition of signals did not significantly alter the mean % mature antigens of the nmap process. Conversely, the mean % mature antigens for the normal file transfer was significantly reduced when the safe signal weight was changed to -2 (all significance assessed through paired t-tests, with 95% confidence demonstrated).

In each experiment the nmap process generated significantly more mature context antigen than any other process. The addition of PAMPs did not significantly increase the detection of the 'anomalous' nmap but combined with a higher safe signal weight, lowered the detection of the normal processes. In future experiments, a much higher level of safe signal could be used without reducing the detection of the misbehaving process (lower rate of false positives).

## 3 Conclusions

In this paper we have demonstrated the use of a Dendritic cell inspired algorithm on a small-scale, real-time problem. The promising results shown in the port scan experiments imply that the DC algorithm plus libtissue framework can be used for the purpose of anomaly detection under real-time conditions. Future work involves the use of a replay client, so data from real-time experiments can be captured for the purpose of testing the parameters and limitations of the system in a controlled manner.

## Acknowledgements

This project is supported by the EPSRC (GR/S47809/01), Hewlet Packard Labs, Bristol, and the Firestorm intrusion detection system team. Libtissue developed by Jamie Twycross.